\documentclass[conference]{IEEEtran}
\IEEEoverridecommandlockouts
\usepackage{cite}
\usepackage{amsmath,amssymb,amsfonts}
\usepackage{algorithmic}
\usepackage{graphicx}
\usepackage{textcomp}
\usepackage{xcolor}
\def\BibTeX{{\rm B\kern-.05em{\sc i\kern-.025em b}\kern-.08em
    T\kern-.1667em\lower.7ex\hbox{E}\kern-.125emX}}

\makeatletter
\newcommand{\linebreakand}{%
  \end{@IEEEauthorhalign}
  \hfill\mbox{}\par
  \mbox{}\hfill\begin{@IEEEauthorhalign}
}
\makeatother

\usepackage{amsmath,amsfonts}
\usepackage{multirow}
\usepackage{multicol}
\usepackage{subcaption}

\usepackage{enumitem}
\usepackage{arydshln}
\usepackage{bm}
\usepackage{amsthm}

\usepackage{color}
\usepackage{algorithm}
\usepackage{algorithmic}
\usepackage{float}
\usepackage{bbding}
\definecolor{orange}{RGB}{255,127,0}

\renewcommand{\Sigma}{\mathfrak{S}}

\def\eqref#1{equation~\ref{#1}}

\def\1{\bm{1}}

\DeclareMathAlphabet{\mathsfit}{\encodingdefault}{\sfdefault}{m}{sl}
\SetMathAlphabet{\mathsfit}{bold}{\encodingdefault}{\sfdefault}{bx}{n}

\def\sR{{\mathbb{R}}}

\def\sZ{{\mathbb{Z}}}

\usepackage{graphicx}
\usepackage{amssymb}
\usepackage{bm}
\usepackage{times}
\usepackage{makecell}
\usepackage{bbm}
\usepackage{xcolor}
\usepackage{xspace}
\usepackage{float}
\usepackage{color}
\usepackage{colortbl}

\usepackage{hyperref}

\usepackage{array}
\newcolumntype{I}{!{\vrule width 3pt}}
\newlength\savedwidth

\newlength\savewidth

\newcommand{\flowtext}{FlowText\xspace}

\def\varnothing{\emptyset}

\renewcommand{\texttt}[1]{ $ {{\tt #1} } $}  
\usepackage{dsfont}

\usepackage[margin=4pt,font=small,labelfont=bf,labelsep=endash,tableposition=top]{caption}

\def\eg{\emph{e.g., }}

\def\ie{\emph{i.e., }}

\definecolor{mygray}{gray}{.92}

\begin{document}


\title{FlowText: Synthesizing Realistic Scene Text Video with Optical Flow Estimation}

\author{\IEEEauthorblockN{Yuzhong Zhao}
\IEEEauthorblockA{\textit{School of Computer Science and Technology} \\
\textit{University of Chinese Academy of Sciences}\\
Beijing, China \\
zhaoyuzhong20@mails.ucas.ac.cn}
\and
\IEEEauthorblockN{Weijia Wu}
\IEEEauthorblockA{\textit{Key Laboratory for Biomedical Engineering of Ministry} \\
\textit{Zhejiang University}\\
Hangzhou, China \\
weijiawu@zju.edu.cn}
\linebreakand 
\IEEEauthorblockN{Zhuang Li}
\IEEEauthorblockA{\textit{MMU} \\
\textit{Kuaishou Technology}\\
Beijing, China \\
lizhuang@kuaishou.com}
\and
\IEEEauthorblockN{Jiahong Li}
\IEEEauthorblockA{\textit{MMU} \\
\textit{Kuaishou Technology}\\
Beijing, China \\
lijiahong@kuaishou.com}
\and
\IEEEauthorblockN{Weiqiang Wang*\thanks{*Corresponding author}}
\IEEEauthorblockA{\textit{School of Computer Science and Technology} \\
\textit{University of Chinese Academy of Sciences}\\
Beijing, China \\
wqwang@ucas.ac.cn}
}

\maketitle

\begin{abstract}
Current video text spotting methods can achieve preferable performance, powered with sufficient labeled training data. 
However, labeling data manually is time-consuming and labor-intensive. To overcome this, using low-cost synthetic data is a promising alternative.
This paper introduces a novel video text synthesis technique called \flowtext, which utilizes optical flow estimation to synthesize a large amount of text video data at a low cost for training robust video text spotters.
Unlike existing methods that focus on image-level synthesis, \flowtext concentrates on synthesizing temporal information of text instances across consecutive frames using optical flow. This temporal information is crucial for accurately tracking and spotting text in video sequences, including text movement, distortion, appearance, disappearance, shelter, and blur.
Experiments show that combining general detectors like TransDETR with the proposed \flowtext produces remarkable results on various datasets, such as ICDAR2015video and ICDAR2013video. Code is available at \url{https://github.com/callsys/FlowText}. 
\end{abstract}

\begin{IEEEkeywords}
Synthetic data, Video text spotting.
\end{IEEEkeywords}

\section{Introduction}
Video text spotting is a task that involves detecting, tracking, and reading text in a video sequence, and has gained popularity due to its various applications in computer vision, such as video understanding~\cite{srivastava2015unsupervised}, video retrieval~\cite{dong2021dual}, video text translation, and license plate recognition~\cite{anagnostopoulos2008license}, etc.
%
Current video text spotters~\cite{wu2022end,wu2022real} achieve preferable performance, powered with sufficient labeled training data.
However, manually annotating such data is time-consuming, expensive, and prone to human errors.
%
According to the annotation report of BOVText~\cite{wu2021bilingual} dataset, $2,021$ videos, $7,292,261$ text instances require the work of $30$ personnel over three months, \textit{i.e.,} $21,600$ man-hours, which is time-consuming and frustrating. 
Moreover, it is also difficult to collect enough data to cover various applications from traffic sign reading to video retrieval tasks.

To reduce the cost of video text annotation and collection, an alternative way is to utilize synthetic data, which is largely available and the ground truth can be freely generated at a low cost.
Previous image-based synthesis algorithms~\cite{zhan2018verisimilar,gupta2016synthetic} have proven they are beneficial and potential for image-level text tasks.
SynthText~\cite{gupta2016synthetic} firstly attempt to propose a synthetic data engine, which overlays synthetic text in existing background images, accounting for the local 3D scene geometry.
Based on SynthText, VISD~\cite{zhan2018verisimilar} tries to synthesize more verisimilar image data by using semantic segmentation and visual saliency to determine the embedding locations.
However, the above synthetic engines only focus on the synthesis quality of image-level. None of them are dedicated to generating effective and efficient video-based content, which is particularly challenging for tasks such as video text spotting.
Compared to image-level synthetic algorithms, video tasks present mainly two challenges. Firstly, video synthesis requires the generation of verisimilar spatiotemporal information, including the movement and deformation of text in a video sequence. This information is vital for spatiotemporal modeling of video text spotting methods, and cannot be provided by image-based synthesis. Secondly, text in video sequences generally presents more complex and challenging cases than static images, due to issues such as motion blur, out-of-focus, artifacts, and occlusion.
To address these challenges, we propose a novel video synthesis technique that incorporates optical flow estimation, which we call \flowtext.


%

%
Our main contributions are summarized as follows:
\begin{itemize}
    \item 
    We propose a new technique for synthesizing video called FlowText, which focuses on creating realistic scene text video, even in complex situations such as motion blur, occlusion, and being out of focus.
    
    \item \flowtext covers a wide range of text scenarios in video sequences, \ie{} motion blur, occlusion, out of focus.
    
    \item As the first video text synthesis method, \flowtext achieves significant improvement compared with other synthesis methods on two datasets for multi-stage tasks~(\ie{}video text detection, tracking, and spotting). Especially, \flowtext achieves $60.1\%$ ${\rm ID_{F1}}$ for video text tracking task and $66.5\%$ ${\rm ID_{F1}}$ for video text spotting task on ICDAR2015video~\cite{karatzas2015icdar}, with $2.0\%$ and $3.7\%$ improvements than the previous SOTA methods, respectively.
    
\end{itemize}

\begin{figure*}[htbp]
 \centering  
 \includegraphics[width=0.98\linewidth]{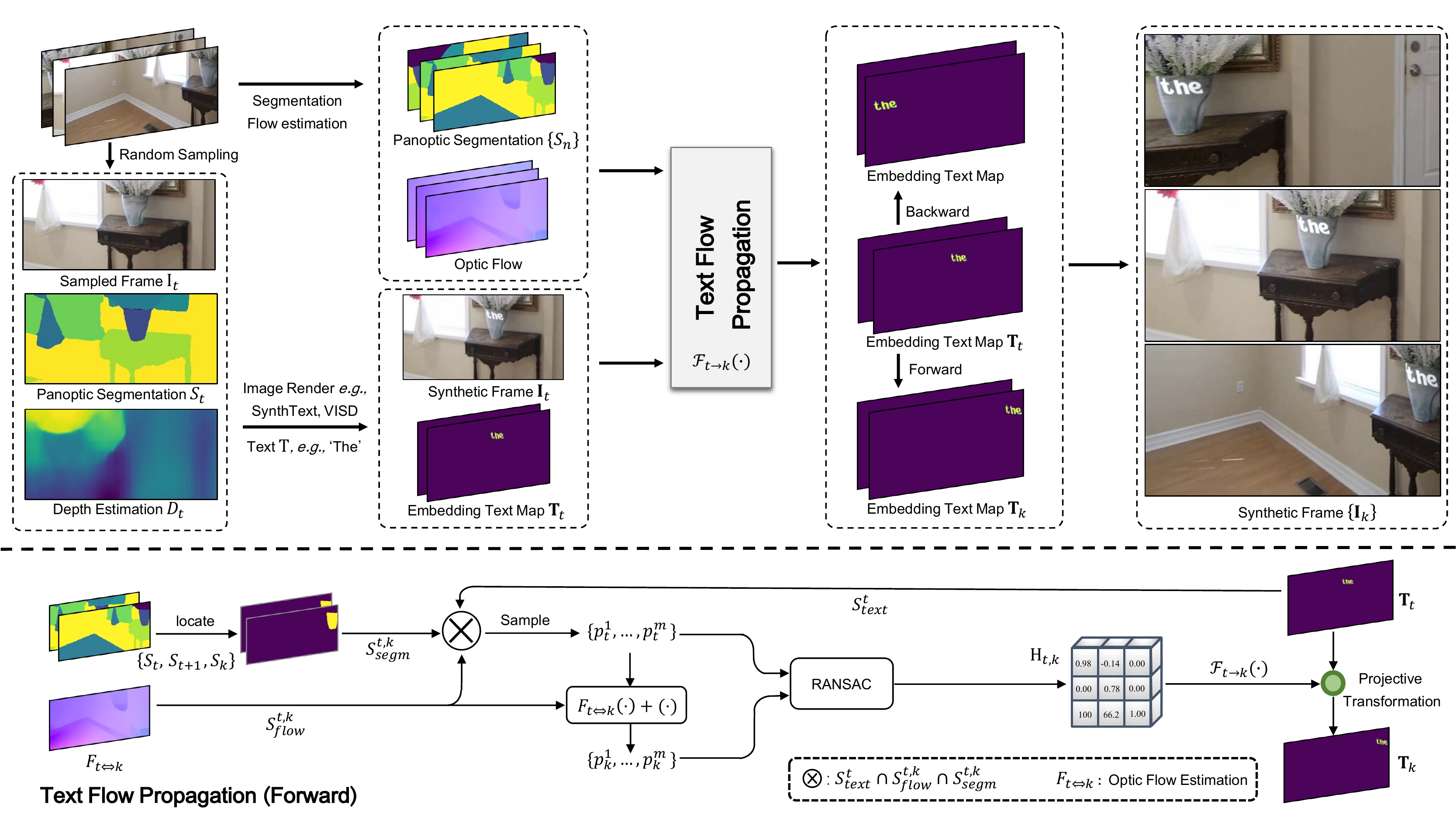}
 \caption{\textbf{Pipeline of the proposed \flowtext}. Upper: For each video, we first randomly sample and render a frame $\text{I}_t$ with an image synthesis method~(\eg{} SynthText, VISD). The Text Flow Propagation~(TFP) is used to propagate the synthetic visual text effects $\textbf{T}_t$ to other frames. Down: The detailed architecture of the TFP.}
\label{fig:pipeline}
\end{figure*}

\section{Related work}

\subsection{Video Text Spotting}
Video scene text spotting~\cite{yin2016text,wu2020texts,wu2019textcohesion} has been studied for years and it has attracted increasing interest in recent years due to its numerous applications in computer vision. 
ICDAR2013 video~\cite{karatzas2013icdar}, the first competition for the task, established the first public dataset to push the field.
ICDAR2023 DSText~\cite{wu2023icdar} presents a new challenge for dense and text scenarios of video text spotting.
As for the algorithm, based on Wang \textit{et al.}~\cite{wang2011end}, Nguyen \textit{et al.}~\cite{nguyen2014video} firstly try to explore and propose an end-to-end solution for text recognition in natural images to video, which trains local character models and explore methods to capitalize on the temporal redundancy of text.
Wang \textit{et al.}~\cite{wang2017end} proposed an multi-frame tracking based video text spotting method, which firstly detects and recognizes text in each frame of the input video, then associates texts in the current frame and several previous frames to obtain the tracking results by post-processing. 
Cheng \textit{et al.}~\cite{cheng2019you} and Cheng \textit{et al.}~\cite{cheng2020free} propose a video text spotting framework by only recognizing the text one-time to replace frame-wisely recognition, which includes a spatial-temporal detector and text recommender to detect and recognize text.
Wu \textit{et al.}~\cite{wu2021bilingual} adopts query concept in the transformer to model text tracking, then uses an attention-based recognizer to obtain the recognition results.
TransDETR~\cite{wu2022end} proposes an end-to-end trainable video text spotting framework with transformer, which each text query is responsible to predict the entire track trajectory and the content of a text in a video sequence.

The methods mentioned above heavily rely on manually annotated images from real-world datasets such as COCOText~\cite{veit2016coco} and ICDAR2015video~\cite{karatzas2015icdar}. However, these datasets are expensive to create and still too small to cover the wide range of text appearances in scenes. 
And previous works~\cite{zhan2018verisimilar,gupta2016synthetic} all have proved the effectiveness and potential of synthetic data, which is a good way to solve the problem.

\subsection{Synthetic Data}
Synthetic data~\cite{gupta2016synthetic,zhan2018verisimilar,wu2023diffumask,siyu2022unsupervised,chen2020stable,chen2022curiosity,wu2020synthetic}, which is generated at a lower cost than manual annotations and detailed ground-truth annotations, has gained increasing attention. In the field of image-level text synthesis, there are some existing synthetic datasets that have become standard practices and are used as pre-trained datasets.
The first work for image text synthesis is SynthText~\cite{gupta2016synthetic}, which blended text into existing natural image scenes using off-the-shelf segmentation and depth estimation techniques to align text with the geometry of the background image and respect scene boundaries. Following SynthText, VISD~\cite{zhan2018verisimilar} proposed three improvements to obtain more verisimilar synthetic images: semantic coherence, better embedding locations with visual saliency, and adaptive text appearance.
However, these methods only focus on image-level synthesis, and there are currently no video-based synthetic engines. There are also challenges in spatiotemporal information synthesis that need to be addressed.

\subsection{Optical Flow}
Optical flow estimation is a fundamental task in computer vision. Classical approaches, such as~\cite{horn1981determining}, typically model optical flow estimation with brightness constancy and spatial smoothness. However, these methods often struggle to handle large displacements and occlusions. In recent years, some approaches have used the coarse-to-fine approach~\cite{hu2016efficient,sun2018pwc} and iterative refinements~\cite{teed2020raft} to handle large displacements incrementally.
RAFT~\cite{teed2020raft} represents the iterative refinement approach and proposes to gradually improve the initial prediction with a large number of iterative refinements, achieving remarkable performance on standard benchmarks. Regarding occlusions, existing approaches~\cite{revaud2015epicflow,xu2017accurate} conduct a forward-backward consistency check to identify and solve occluded regions with interpolation.
GMA~\cite{jiang2021learning} is the first work to take an implicit approach to the occlusion challenge. It adopts global motion features to predict flow accurately in occluded regions. In video data synthesis, optical flow plays a vital role in ensuring smooth and verisimilar temporal text synthesis. Compared to large displacements, occlusions pose a more realistic challenge in video synthesis. Therefore, we adopt GMA~\cite{jiang2021learning} as the base optical flow estimation model for our \flowtext.


\section{Method}
\subsection{Formulation}

While generating synthetic video automatically demonstrate great potential in the field of video text, producing high-quality synthetic video remains a significant challenge. One possible solution to this challenge is to generate a single frame using an image synthesis method, and then iteratively map the synthetic visual effects of embedding text to adjacent frames using spatiotemporal relevance information in the video.

We take the pasting process of a single text $\text{T}$ as an example (\eg{}``\texttt{The}'' in Fig.~\ref{fig:pipeline}).
Given one video sequence $\{\text{I}_k\}_{k\in\mathcal{N}}, \text{I}_k\in\sR^{h\times w}$, $\mathcal{N}=\{1,2,\ldots,n\}$, where $h$, $w$, and $n$ are the height, width, and length of the video. 
As shown in Fig.~\ref{fig:pipeline}, we randomly sample the $t$-th frame $\text{I}_t$ from the video.
%
Then, the corresponding synthetic image $\mathbf{I}_t\in\sR^{h\times w}$ and the embedding text map $\mathbf{T}_t$  (representing visual text feature of text $\text{T}$, \eg{}font and shape) of the sampled frame can be obtained with the existing image synthesis method (\eg{}SynthText~\cite{gupta2016synthetic}).
Next, a spatiotemporal propagation function $\mathcal{F}_{t \to k}(\cdot )$ is calculated for mapping the embedding text map $\mathbf{T}_t$ to that of other frames (\eg{}$\mathbf{T}_{k}$).
Finally, the whole synthetic text video $\{\mathbf{I}_k\}_{k\in\mathcal{N}}$ is obtained by embedding the corresponding embedding text maps $\{\mathbf{T}_k\}_{k\in\mathcal{N}}$ to the video sequence $\{\text{I}_k\}_{k\in\mathcal{N}}$. 
%
%
However, obtaining the propagation function $\mathcal{F}_{t \to k}(\cdot )$ is challenging. To solve this problem, we propose the Text Flow Propagation (TFP) Algorithm, which uses optical flow to fit the function.

\subsection{Overview of FlowText}
In this section, we will provide a comprehensive introduction of the entire \flowtext pipeline, which consists of two main steps: \textit{Rendering Sampled Frame} and \textit{Text Flow Propagation}.

\subsubsection{Rendering Sampled Frame} as shown in Fig.~\ref{fig:pipeline} (upper), for each video sequence, we firstly randomly select the $t$-th frame $\text{I}_t$ as the sampled frame.
Then, we adopt the approach used in SynthText~\cite{gupta2016synthetic} and VISD~\cite{zhan2018verisimilar} to overlay the text onto the image. The location and orientation of the text are determined by the depth map $D_t$ predicted by Monodepth2~\cite{godard2019digging} and the panoptic segmentation map $S_t$ predicted by Mask2former~\cite{cheng2022masked}, respectively.
Especially, it can be formulated as:
\begin{align}
    \left \{ \mathbf{I}_t,\mathbf{T}_t  \right \}  =  \underset{\mathrm{Render}}{\mathbb{M}}  (\text{I}_t,\text{T})|_{S_t,D_t}\,,
\end{align}
where $\mathbf{I}_t$ and $\mathbf{T}_t$ denote synthetic frame and embedding text map of text $\text{T}$ in the $t$-th frame.
$\underset{\mathrm{Render}}{\mathbb{M}}(\cdot ) $ refer to SynthText~\cite{gupta2016synthetic} in this paper.

\subsubsection{Text Flow Propagation} After obtaining the embedding text map $\mathbf{T}_t$, we can calculate the $\mathbf{T}_{k}$ with the Text Flow Propagation algorithm $\mathcal{F}_{t \to k}(\cdot )$ as:
%

\begin{align}
    \mathbf{T}_{k} = \mathcal{F}_{t \to k}(\mathbf{T}_t)\,.
\end{align}
%
Finally, we produce the whole synthetic text video $\{\mathbf{I}_k\}_{k\in\mathcal{N}}$ via overlapping the embedding text maps $\{\mathbf{T}_k\}_{k\in\mathcal{N}}$ to the video sequence $\{\text{I}_k\}_{k\in\mathcal{N}}$.

\subsection{Text Flow Propagation Algorithm}
There are two algorithms for Text Flow Propagation (TFP): Forward Text Flow Propagation (FTFP, $\mathcal{F}_{t \to k},k>t$) and Backward Text Flow Propagation (BTFP, $\mathcal{F}_{t \to k},k<t$), depending on whether the estimated frame is located before or after the sampled frame. Despite the difference in optical flow direction, both algorithms are quite similar.
For convenience, we take FTFP as the example. FTFP is based on an existing optical flow estimation model (GMA~\cite{jiang2021learning} in this paper), which can be directly used to map points between frames as:
\begin{align}
    p_{k} = F_{t\Leftrightarrow k}(p_t) +p_t,\quad p_t\in \textbf{T}_t, p_k\in \textbf{T}_k, 
    \label{denseflow}
\end{align}
where $p_k$ represents the coordinate of point $p$ at $k$-th frame. However, there are main two obvious problems that affect the performance of the optical flow estimation: (1) \textit{Unconstrained mapping}: Optical flow estimation does not view text geometry as a whole, and destroy the invariant properties, \eg{}concurrency, collinearity, order of contact. (2) \textit{Error mapping due to occlusion and noise}: Occlusion and outliers will cause inaccurate mapping and inauthentic synthetic data.

In this paper, we propose to solve the first problem by \textit{Projective Transformation} and solve the second problem by \textit{Point Resample}.

\begin{figure}[tbp]
	\begin{minipage}{0.98\linewidth}
	\includegraphics[width=0.99\linewidth]{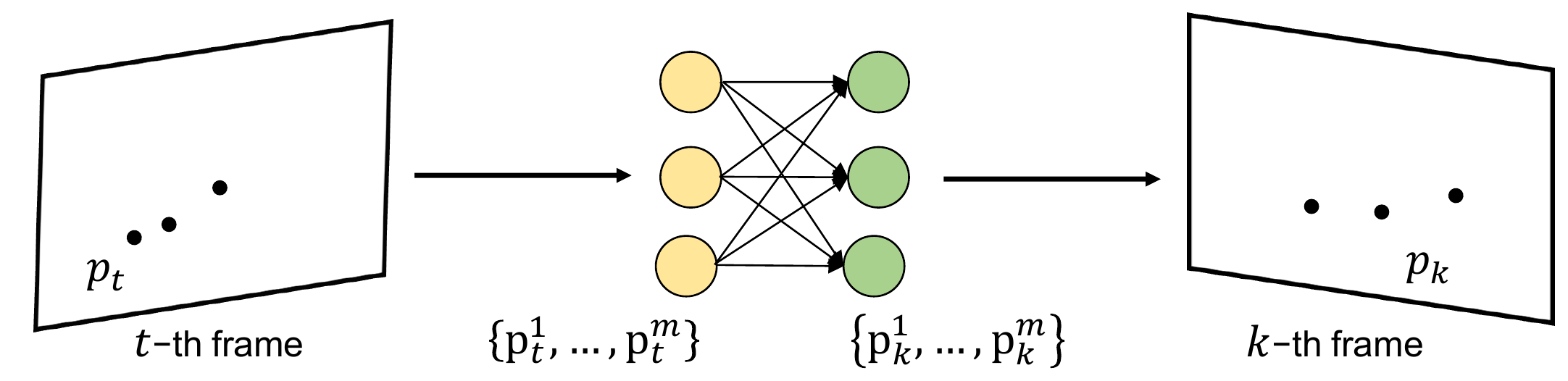}
	\subcaption{Mapping with Dense Optical Flow Estimation.}
	\label{fig:1a}	
	\end{minipage}
	\qquad
	\begin{minipage}{0.98\linewidth}
	\centering
	\includegraphics[width=0.99\linewidth]{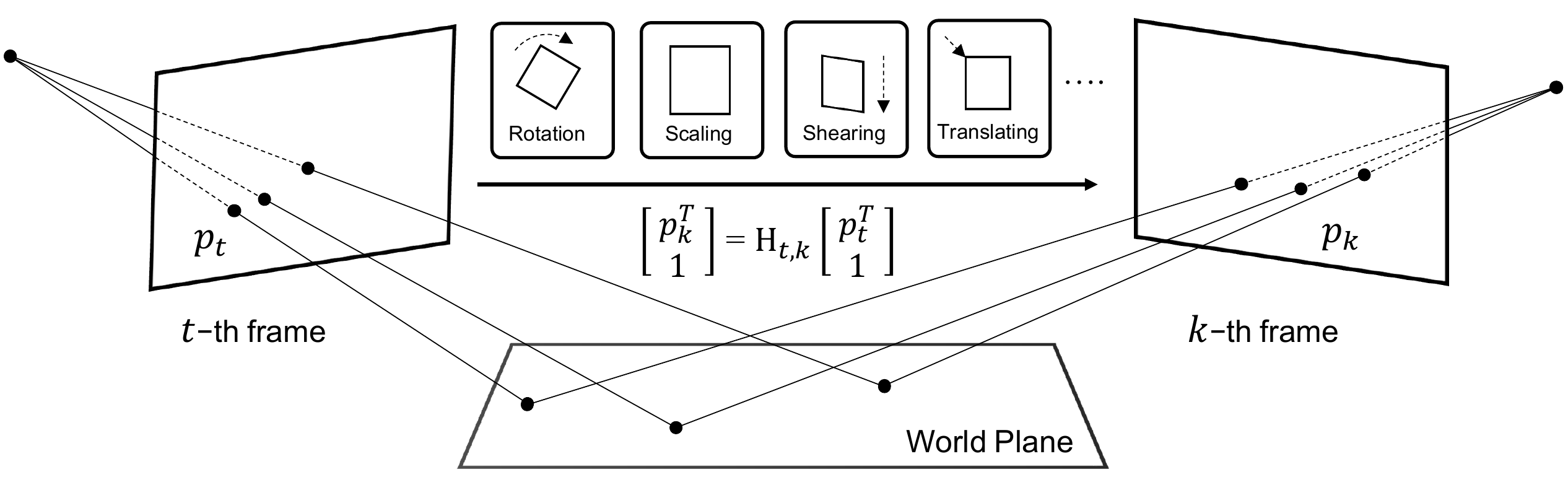}
	\subcaption{Mapping with Projective Transformation (Ours).}
		 \label{fig:1b}	
	\end{minipage}
	\caption{\textbf{Comparison of different mapping methods.} Mapping with projective transformation present better multiple view geometry transformation via view the embedding text map $\textbf{T}$ as a whole.
}
\label{fig:motivation}
\end{figure}

\subsubsection{Mapping with Projective Transformation} 
%
We view the mapping function as a multiple view geometry problem, where text is painted on a planar surface~(\eg{}wall or sign).
The same text in different frames is observed in different view planes, as shown in Fig.~\ref{fig:1b}.
And they can be transformed into each other with projective transform~\cite{hartley2003multiple}.
Specially, with a $3 \times 3$ homography matrix, \ie{}$\rm{H}_{t,k}\in \sR^{3\times 3}$, we can formulate the mapping function $\mathcal{F}_{t \to k}$ as:
\begin{equation}
\begin{aligned}
\left[\begin{array}{c} {p_k}^T\\ 1 \end{array}\right]&= \rm{H_{t,k}^{-1}} 
\left[\begin{array}{c} {p_t}^T \\ 1 \end{array}\right],\quad p_t\in \textbf{T}_t, p_k\in \textbf{T}_k, 
\label{eq:trans}
\end{aligned}
\end{equation}

where the homography matrix $\rm{H}_{t,k}$ transforms the coordinates of the embedding text map $\textbf{T}_t$ to $\textbf{T}_k$.   
Compared with dense optical flow mapping (Fig.~\ref{fig:1a}), the projective transform~\cite{hartley2003multiple} (Fig.~\ref{fig:1b}) keeps invariant properties of text (\ie{}concurrency, collinearity, order of contact). 
And its degree of freedom~(df) is $8$ (\eg{}rotation, scaling, shearing, translating), which makes the mapping function stable.

According to the book `Multiple View Geometry'~\cite{hartley2003multiple}, the exact solution for the matrix $\rm{H}_{t,k}$ is possible, if the below theorem can be ensured.

\noindent\textbf{Theorem 1.} \textit{The 2D projective transformation:  $\mathrm {P^2}\to \mathrm {P^2}$ can be determined if there exists at lease four correspondences to calculate a non-singular $3 \times 3$ matrix $\rm{H}_{t,k}$.}

To estimate $\rm{H}_{t,k}$, we first identify points inside the text region in the $t$-th frame, denoted as $S_{text}^t=\{p_t\mid \mathbf{T}_t(p_t)>0, p_t\in\sZ^2\}~\label{eq:stext}$. Here, $p_t$ is a 2D point coordinate and $\mathbf{T}_t(p_t)$ is the value of $\mathbf{T}_t$ at point $p_t$. Next, we map each point $p_t$ to its corresponding point $p_k$ in the $k$-th frame using Equ.~\ref{denseflow}. We then collect the point pairs ${(p_t, p_k)}$, where $p_t\in S_{text}^t$. Finally, we estimate the projective matrix $\rm{H}_{t,k}$ by using the RANdom SAmple Consensus (RANSAC) algorithm~\cite{fischler1981random} to robustly fit the point pairs.

\begin{figure}[tbp]
 \centering  
 \includegraphics[width=0.98\linewidth]{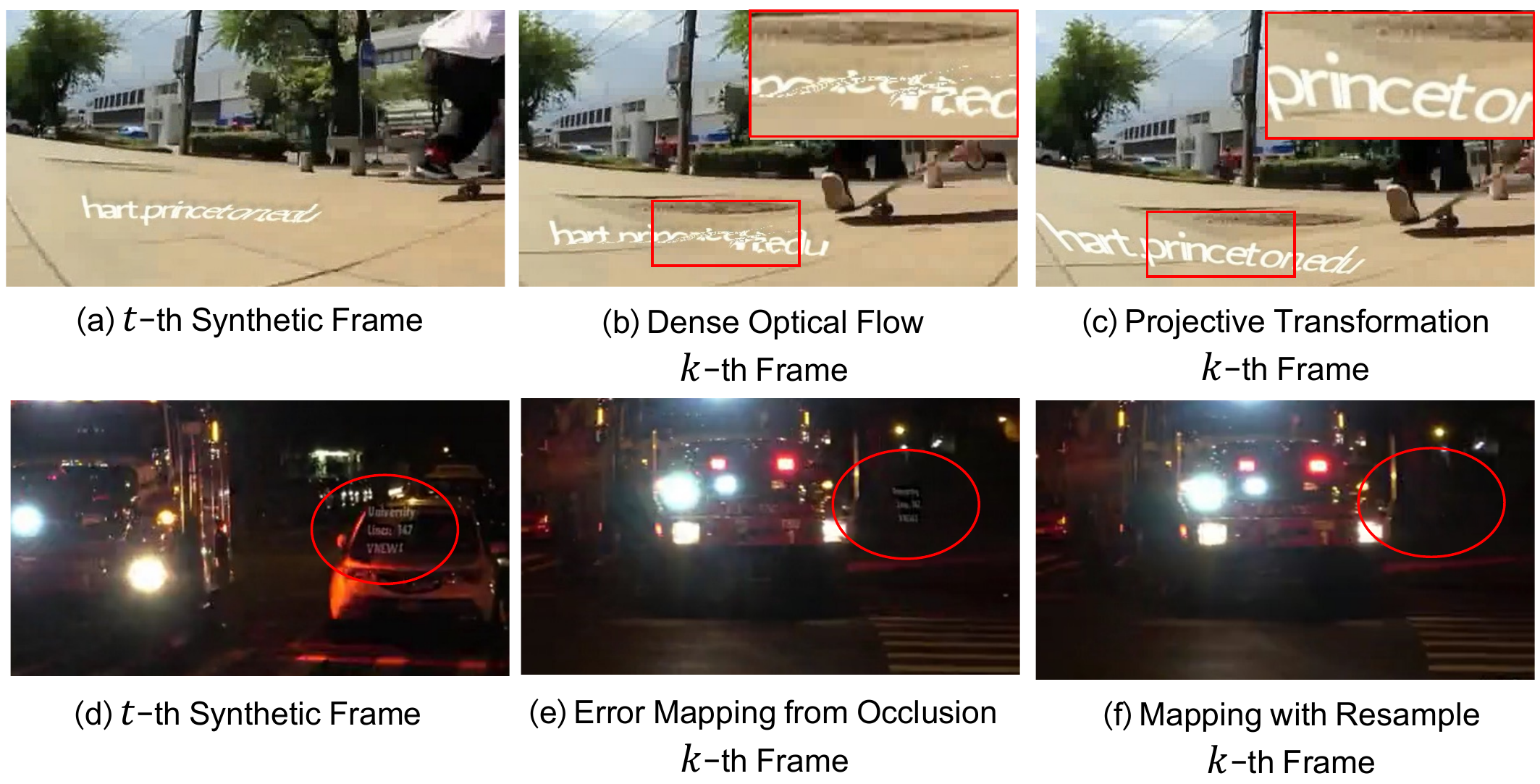}
 \caption{\textbf{Illustration of the two main points in our TFP}. Upper: Projective transformation presents better performance than dense optical flow mapping. Down: Resample with two constrains can refine the error from occlusion.}
\label{fig:case1}
\end{figure}

\subsubsection{Point Resample} 
\label{332}

When occlusion occurs, abnormal sampled coordinate points in $S_{text}^t$ can lead to unreasonable mapping associations and result in unreal text distortion. To address this issue, we can use optical flow and segmentation constraints to remove these abnormal points.
In dense flow estimation, some points can be outliers or noise and deviate from other sampled points. We can use statistics, specifically the standard deviation ($\sigma$) and mean ($\mu$) of the L2 normalization of point pairs ($||F_{t\Leftrightarrow k}(p_t)||_2, p_t\in S_{text}^t$), to detect these outliers. We set the lower limit to ($\mu-\sigma$) and the upper limit to ($\mu+\sigma$). Any sample that falls outside this range is detected as an outlier and removed from the sampled coordinate points $S_{text}^t$. We can define the resampled point set with optical flow constraint as follows:

\begin{align}
    S_{flow}^{t,k}=\{p_t\mid \mu-\sigma\leq||F_{t\Leftrightarrow k}(p_t)||_2\leq\mu+\sigma\}. 
    \label{eq:sflow}
\end{align}

Occlusion often leads to a breakdown in the semantic coherence of text within a video sequence. It is essential that each text is consistently associated with the same semantic entity in all frames of the video (for example, the word ``\texttt{The}" should always be superimposed on the ``\texttt{flowerpot}" in Fig.~\ref{fig:pipeline}).
To achieve this consistency, it is necessary to exclude sampled points that move outside of the semantic entity. We define the resampled point set that adheres to segmentation constraints as follows:
\begin{align}
    S_{segm}^{t,k}=\{p_t\mid S_i(p_i)>0,i=t,t+1,\ldots,k\}, 
    \label{eq:ssegm}
\end{align}
where $S_i$ denotes the segmentation map of the semantic entity in the $i$-th frame.
Finally, we define sampling point set as:
\begin{align}
  S^{t,k} = S_{text}^t \bigcap S_{flow}^{t,k} \bigcap S_{segm}^{t,k}\,.
\label{intersection}
\end{align}

We use the sampling points in $S^{t,k}$ to estimate the projective matrix $\rm{H}_{t,k}$ between the $t$-th frame and the $k$-th frame.


\begin{algorithm}[htbp]
\small
\caption{Forward Text Flow Propagation}
\label{code:ftfp}
\raggedright
\textbf{Input}: \\
\qquad $\mathbf{T}_t$ : the embedding text map of the text in $t$-th frame.\\
\qquad $F_{t\Leftrightarrow k}$ : the optic flow between $t$-th frame and $k$-th frame.\\
\qquad $\{S_i\}_{i=t,t+1,\ldots,k}$ : the segmentation map of the semantic entity between $t$-th frame and $k$-th frame.\\
\textbf{Parameter}: \\
\qquad $N$ : The minimum number of sampling points.\\
\textbf{Output}: \\
\qquad $\mathbf{T}_k$ : the embedding text map in the $k$-th frame.

\begin{algorithmic}[1] 
\STATE Calculate $S_{text}^t$ with Equ.~\ref{eq:stext};
\STATE Calculate $S_{flow}^{t,k}$ with Equ.~\ref{eq:sflow};
\STATE Calculate $S_{segm}^{t,k}$ with Equ.~\ref{eq:ssegm};
\STATE $S^{t,k} \leftarrow S_{text}^{t,k}\bigcap S_{flow}^{t,k}\bigcap S_{segm}^{t,k}$;
\IF{$|S^{t,k}|\leq N$}
\STATE $\mathbf{T}_k = 0^{h\times w}$;
\ELSE

\STATE $\mathcal{P} \leftarrow \varnothing$;
\FOR{$p_t \in S^{t,k}$}
\STATE $p_k = F_{t\Leftrightarrow k}(p_t) + p_t$;
\STATE $\mathcal{P} = \mathcal{P}\cup (p_t,p_k)$;
\ENDFOR

\STATE $\text{H}_{t,k} = \text{RANSAC}(\mathcal{P})$;
\STATE $\mathbf{T}_k = \text{ProjectiveTransform}(\mathbf{T}_t,\text{H}_{t,k})$;
\STATE $\mathbf{T}_k = \mathbf{T}_k \cdot S_k$;
\STATE $\mathbf{T}_k = \text{MotionBlur}(\mathbf{T}_k, F_{t\Leftrightarrow k})$;
\ENDIF
\RETURN $\mathbf{T}_k$;
\end{algorithmic}
\label{alg:tfp}
\end{algorithm}

\begin{table*}[htbp]
    \centering
    \small 
    \setlength{\tabcolsep}{1mm}
    \caption{End to End Video Text Spotting performance on ICDAR2015 video~\cite{nguyen2014video}. FT denotes finetune with real data. `M-Tracked' and `M-Lost' denote `Mostly Tracked' and `Mostly Lost', respectively.}
\begin{tabular}{l|c|c|ccccc}
\hline
\setlength{\tabcolsep}{0.7mm}
\multirow{2}{*}{Data} &  \multirow{2}{*}{FT}&
\multirow{2}{*}{Data Size~(Image) } 
& \multicolumn{5}{c}{ ICDAR2015 video/\%}\\
 &  &   & ${\rm ID_{F1}}$$\uparrow$ & ${\rm MOTA}$$\uparrow$ & ${\rm MOTP}$$\uparrow$ & ${\rm M\mbox{-}Tracked}$$\uparrow$ & ${\rm M\mbox{-}Lost}$$\downarrow$ \\
\hline

SynthText~\cite{gupta2016synthetic} & $\times$ & 800k(Image) &  44.8 &16.6  &70.3 & 20.1 & 57.9   \\
VISD~\cite{zhan2018verisimilar} & $\times$ & 10k(Image) & 44.9 & 22.5 & 70.3 & 17.9 & 58.6   \\
\flowtext(Ours) & $\times$ & 250k~(5k videos) & 48.5 & 25.2 & 73.1 & 24.1 & 49.2   \\
\hline
None & $\checkmark$ & 13k(ICDAR 2015 video) & 59.3  & 44.7 & 74.2 & 32.2 & 43.1   \\
SynthText~\cite{gupta2016synthetic} & $\checkmark$ & 800k(Image)+13k(ICDAR 2015 video)& 62.8  & 50.5 & 74.2 & 35.3 & 39.5   \\
VISD~\cite{zhan2018verisimilar} &$\checkmark$ & 10k(Image)+13k(ICDAR 2015 video)& 60.7 & 45.8 & 74.3 & 33.1 & 43.2   \\
\flowtext(Ours) &$\checkmark$ &250k~(5k videos) + 13k(ICDAR 2015 video) &  66.5 & 52.4 & 74.4 & 38.5 & 35.7   \\
\hline

\end{tabular}
    \label{tab:icdar14}
\end{table*}

\subsubsection{Estimating Motion blur with Optic flow} 
In order to simulate the text motion blur, we add motion blur to the embedding text map $\{\textbf{T}_k\}_{k\in\mathcal{N}}$ according to the direction $\vec{v}$ and scale $||\vec{v}||$ of the optical flow predicted by GMA~\cite{jiang2021learning} in the text region. Motion blur is realized by convolving the embedding text map $\mathbf{T}_k$ with a specific convolution kernel. The convolution kernel apply average pooling for $\alpha||\vec{v}||$ pixels along the direction of $\vec{v}$, where $\alpha$ is a hyperparameter that related to degree of blur.

\subsubsection{Pseudo Code}
We describe the whole process of the Forward Text Flow Propagation in Algorithm~\ref{alg:tfp}. $\text{RANSAC}(\cdot )$ denotes the RANdom SAmple Consensus (RANSAC) algorithm and $\text{ProjectiveTransform}(\cdot )$ denotes transforming the image with the given homography matrix. For the case that Backward Text Flow Propagation (BTFP) is needed, we directly reverse the optical flow and apply FTFP for calculation.

\section{Experiments}

\subsection{Settings}
Following the setting of previous works~\cite{gupta2016synthetic,zhan2018verisimilar}, we verify the effectiveness of the proposed FlowText by training video text spotter on the synthesized images and evaluating them on real image datasets.
In all experiments, we train the model
with 8 Tesla V100 GPUs. The detailed setting of methods all follows the original paper and official code.

\textbf{Benchmark Datasets.} ICDAR2013video  \cite{karatzas2013icdar} is proposed in the ICDAR2013 Robust Reading Competition, which contains 13 videos for
training and 15 videos for testing. These videos are harvested from indoors and outdoors scenarios, and each text is labeled as a quadrangle with 4 vertexes in word-level. 
ICDAR2015video~\cite{karatzas2015icdar} is the expanded version of ICDAR2013 video, which consists of a training set of 25 videos~(13,450 frames) and a test set of 24 video~(14,374 frames). Similar to ICDAR2013 video, text instances in this dataset are labeled at the word level. Quadrilateral bounding boxes and transcriptions are provided. 

\textbf{Text and Video Sources.}
To better simulate the motion of text in video, we use Activitynet~\cite{caba2015activitynet} as the video sources to build FlowText, which contains plenty of complex movement scenarios. For videos in Activitynet, we first use the Kuaishou VideoOCR api~\cite{KuaiShou_api} to filter videos that do not contain text as candidate videos. Then, we random extracte candidate texts from the Newsgroup20 dataset~\cite{newsgroup20} and paint them onto candidate videos with FlowText.

\textbf{Video Text Methods.} 
We use TransDETR~\cite{wu2022end} to evaluate the performance of different synthetic datasets in this paper, which is the State-of-the-art method in both video text tracking and video text spotting domain.

\begin{table}[tbp]
    \centering
    \small 
    \setlength{\tabcolsep}{1mm}
    \caption{Text tracking performance on ICDAR2013 video~\cite{karatzas2013icdar} and ICDAR2015 video~\cite{nguyen2014video}.}
\begin{tabular}{l|c|ccc|ccc}
\hline
\setlength{\tabcolsep}{0.5mm}
\multirow{2}{*}{Data} & \multirow{2}{*}{FT} & \multicolumn{3}{c|}{ICDAR2013 video/\%}& \multicolumn{3}{c}{ICDAR2015 video/\%} \\
 &  &  ${\rm ID_{F1}}$ & ${\rm MOTA}$ &  ${\rm MOTP}$
 & ${\rm ID_{F1}}$ & ${\rm MOTA}$ & ${\rm MOTP}$  \\
\hline

SynthText & $\times$ & 42.9 &  17.3 &  69.8 & 38.2 & 15.9 & 70.4   \\

VISD & $\times$ & 44.7 & 21.4  &  69.6 & 38.0 & 18.0  & 70.2 \\


\flowtext & $\times$ & 47.9 & 27.4 & 74.1 & 41.5 & 21.1 & 72.8  \\
\hline

None & $\checkmark$ & 58.2 & 43.6 & 76.3 &  56.2& 38.7 & 73.0  \\
SynthText & $\checkmark$ & 60.9 & 46.2 & 76.4 & 58.1 & 39.2 &73.1   \\
VISD &$\checkmark$  & 59.4 & 44.7 & 76.4 & 57.7 & 39.8 & 73.1 \\
\flowtext &$\checkmark$ & 64.1 & 48.9 & 76.5 & 60.1 & 42.4 & 73.5 \\
\hline

\end{tabular}




    \label{tab:icdar13}
\end{table}


\subsection{Video Text Tracking}
As shown in Table~\ref{tab:icdar13}, without real data, FlowText outperforms the previous SOTA method by 3.3\% in ${\rm ID_{F1}}$ on ICDAR2015. When real data is introduced, FlowText outperforms the previous SOTA method by 2.0\% on ICDAR2015. This proves that the temporal information simulated by FlowText can effectively improve the tracking performance.

\subsection{End to End Video Text Spotting}
As shown in Table~\ref{tab:icdar14}, with real data, FlowText outperforms the previous SOTA method SynthText by 3.7\% in ${\rm ID_{F1}}$ on ICDAR2015. This proves that FlowText can greatly boost the training of video text spotter.

\begin{table}[tbp]
    \centering
    \small 
    \setlength{\tabcolsep}{1mm}
    \caption{Ablation experiments on ICDAR2013 video.}
\begin{tabular}{cccc|ccc}
\hline
\setlength{\tabcolsep}{0.5mm}
\multirow{2}{*}{Flow} & \multirow{2}{*}{Projective} &\multirow{2}{*}{Resample} &\multirow{2}{*}{Blur} & \multicolumn{3}{c}{ICDAR2013 video/\%}
\\
 &  & & & ${\rm ID_{F1}}$ &${\rm MOTA}$ &  ${\rm MOTP}$ \\

\hline

& & & & 36.7 & 19.0 & 72.0  \\

\checkmark & & & & 39.1 & 19.9 & 71.0 \\

\checkmark & \checkmark& & & 40.9 & 24.1 & 73.0 \\

\checkmark & \checkmark& \checkmark& & 47.5 & 27.0 & 74.0 \\

\checkmark & \checkmark& \checkmark& \checkmark  & 47.9 & 27.4 & 74.1 \\

\hline
\end{tabular}
    \label{tab:ablation}
\end{table}

\begin{table}[tbp]
    \centering
    \small 
    \setlength{\tabcolsep}{1mm}
    \caption{Ablation of video source. We maintain that we generate 5k videos and 10 frame per video for the experiment.}
\begin{tabular}{c|ccc}
\hline
\setlength{\tabcolsep}{0.5mm}
\multirow{2}{*}{Video source} & \multicolumn{3}{c}{ICDAR2013 video/\%}
\\
 & ${\rm ID_{F1}}$ &${\rm MOTA}$ &  ${\rm MOTP}$ \\

\hline

Youcook2~\cite{ZhXuCoCVPR18} & 37.7 & 19.6 & 72.5 \\

GoT10k~\cite{huang2019got} & 36.9 & 21.0 & 71.5 \\

Activitynet~\cite{caba2015activitynet} & 44.6 & 26.1 & 72.8 \\

\hline
\end{tabular}
    \label{tab:ablation4}
\end{table}

\subsection{Ablation Study}
\subsubsection{Main components}As shown in line 1 of Table~\ref{tab:ablation}, the image render used in FlowText (\ie{SynthText}) achieve 36.7\% ${\rm ID_{F1}}$ on ICDAR2013. When we directly use optic flow to propagate temporal information for text, relative improvement of 2.4\% in ${\rm ID_{F1}}$ can be achieved. Then we use projective transform to replace optic flow wrapping, which brings relative improvement of 1.8\% in ${\rm ID_{F1}}$. Finally, we use resample to remove abnormal sampled points in TFP and apply motion blur to the text map, which brings relative improvement of 6.6\% and 0.4\% in ${\rm ID_{F1}}$ respectively. The ablation experiment proves that introducing optical flow information into synthetic data can improve the performance of video text spotter, and the proposed TFP algorithm can effectively improve the quality of the synthesized text.

\subsubsection{Video sources} As shown in Table~\ref{tab:ablation4}, we test three datasets as video sources, and Activitynet~\cite{caba2015activitynet} get the best performance. Actvitynet is built for action recognition, which contains plenty of complex movement scenarios. Synthesis videos generated based on Activitynet can obtain more spatiotemporal information.

\section{Conclusion}
In this paper, we propose a novel video synthesis technique called FlowText, which generate fully labeled scene text video with no annotation cost. As the first video text synthesis method, FlowText achieves significant improvement compared with other synthesis methods for multi-stage tasks (\ie{video text detection, tracking and spotting}).

\bibliographystyle{Bib/IEEEtran}
\bibliography{flowtext}  

\end{document}